\documentclass[letterpaper]{article} 
\usepackage[submission]{aaai2026}  
\usepackage{times}  
\usepackage{helvet}  
\usepackage{courier}  
\usepackage[hyphens]{url}  
\usepackage{graphicx} 
\def\UrlFont{\rm}  
\usepackage{natbib}  
\usepackage{caption} 
\usepackage{algorithm}
\usepackage{algorithmic}

\usepackage{newfloat}
\usepackage[colorinlistoftodos,prependcaption]{todonotes}
\usepackage{xr}
\usepackage{subcaption}
\usepackage{fancyvrb}

\usepackage{listings}
\DeclareCaptionStyle{ruled}{labelfont=normalfont,labelsep=colon,strut=off} 
\floatstyle{ruled}
\newfloat{listing}{tb}{lst}{}
\floatname{listing}{Listing}
\title{Active Query Selection for Crowd-Based Reinforcement Learning}
\author{
    Jonathan Erskine\equalcontrib\thanks{Jonathan Erskine's PhD research is jointly funded by UK Research and Innovation (UKRI) and Thales Training \& Simulation Ltd. through the UKRI Centre for Doctoral Training in Interactive Artificial Intelligence under grant EP/S022937/1. This work was partially funded by the UKRI Turing AI Fellowship EP/V024817/1.},
    Taku Yamagata\equalcontrib,
    Raúl Santos-Rodgríguez
}
\affiliations{
    University of Bristol\\
    \{jonathan.erskine,taku.yamagata,enrsr\}@bristol.ac.uk

}

\usepackage{bibentry}

\usepackage{amsmath, amssymb}
\usepackage[nolist]{acronym}
\usepackage{booktabs}
\usepackage{enumitem}
\usepackage{colortbl} 

\begin{document}

\maketitle

\begin{abstract}

Preference-based reinforcement learning has gained prominence as a strategy for training agents in environments where the reward signal is difficult to specify or misaligned with human intent. However, its effectiveness is often limited by the high cost and low availability of reliable human input, especially in domains where expert feedback is scarce or errors are costly. To address this, we propose a novel framework that combines two complementary strategies: probabilistic crowd modelling to handle noisy, multi-annotator feedback, and active learning to prioritize feedback on the most informative agent actions. We extend the Advise algorithm to support multiple trainers, estimate their reliability online, and incorporate entropy-based query selection to guide feedback requests. We evaluate our approach in a set of environments that span both synthetic and real-world-inspired settings, including 2D games (Taxi, Pacman, Frozen Lake) and a blood glucose control task for Type 1 Diabetes using the clinically approved UVA/Padova simulator. Our preliminary results demonstrate that agents trained with feedback on uncertain trajectories exhibit faster learning in most tasks, and we outperform the baselines for the blood glucose control task.
\end{abstract}

\section{Introduction}
Preference-based reinforcement learning has become a powerful paradigm for tailoring the behaviour of autonomous agents with human trainers. By incorporating human judgments, typically in the form of pairwise comparisons indicating whether a particular action is preferred over the other, human preferences can guide exploration, improve sample efficiency, and prevent unsafe behaviours during training \cite{kaufmann2024survey}. However, practical deployment remains limited by two central challenges: (1) the cost of collecting high-quality feedback from crowds of humans with mixed skills and motives and (2) the limited availability of reliable human trainers, particularly in domains where expertise is scarce or errors carry high risk (e.g., healthcare, robotics).

One approach to mitigate the scarcity of expert feedback is to aggregate input from multiple non-expert trainers, a technique broadly studied in the learning from crowds literature. These methods aim to infer both the true label and trainer reliability by modelling the noise inherent in human inputs. However, while crowd-based strategies can scale feedback collection \cite{buecheler2010}, not all trainers are equally reliable.

A second, complementary strategy to reduce annotation burden is active learning. In supervised settings, active learning algorithms prioritize annotation of the most uncertain or informative data points to accelerate training with fewer labels. In RLHF, however, relatively little work has been done to adapt such ideas to interactive feedback-driven learning. Yet, not all agent behaviours are equally valuable to label; querying human feedback only on high-uncertainty actions could yield greater improvements in policy quality for the same feedback budget.

Many recent works collect feedback as pairwise comparisons of agent actions or trajectories \cite{Abdelkareem_2022} and tend to convert these comparisons into a reward signal \cite{chakraborty2024maxminrlhf} and/or generate models which can learn to predict these preferences \cite{frick2024evaluaterewardmodelsrlhf}. In our experiments we instead prescribe to the practice of collecting feedback directly on the policy as an additional signal to environmental rewards, as demonstrated by \citet{Griffith2013}. 

In this work, we propose a novel feedback aggregation framework that combines active learning, learning from crowds, and policy feedback: probabilistic crowd modelling to handle noisy, multi-trainer feedback, and entropy-based active selection of state-action pairs to maximize the utility of human input. Building on the framework provided in \cite{Griffith2013}, our method aims to:

\begin{enumerate}
    \item Handle feedback from multiple trainers of varying reliability.
    \item Infer trainer consistency levels in an online setting.
    \item Actively request feedback only for agent behaviours deemed highly uncertain.
\end{enumerate}

\begin{figure*}[t]
\centering
\includegraphics[width=0.95\textwidth]{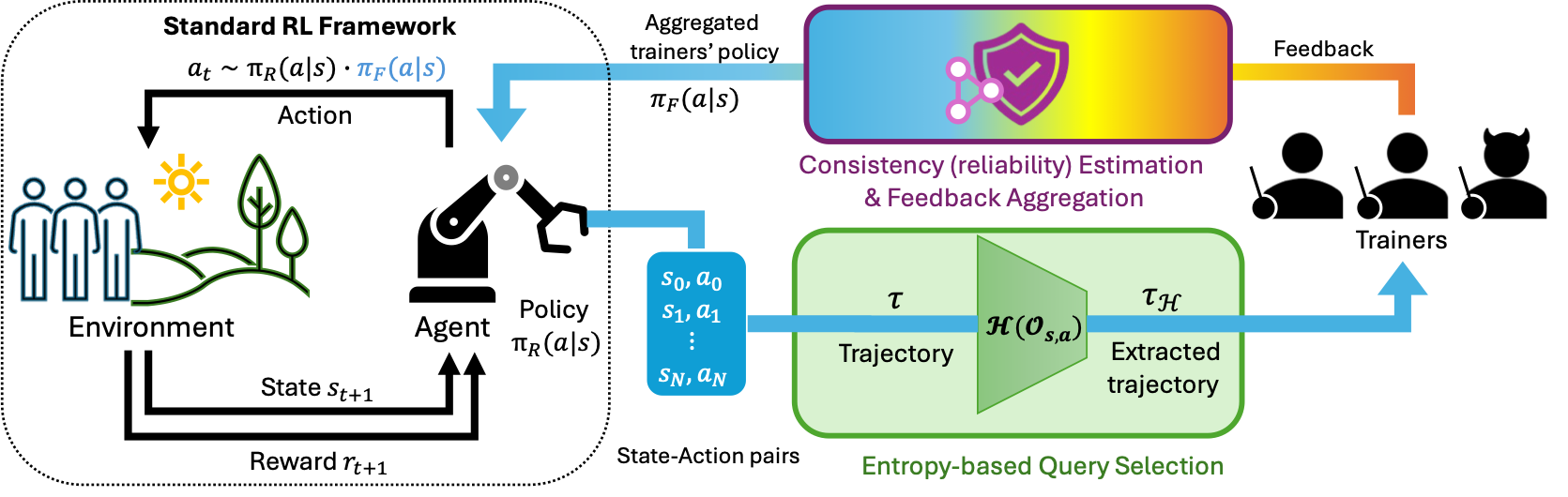} 
\caption{Standard reinforcement learning training loop and our extended method. We collect a trajectory \(\boldsymbol{\tau}\) of state–action pairs and extract a subset \(\boldsymbol{\tau}_{\mathcal{H}} \subseteq \boldsymbol{\tau}\) consisting of the top-\(n\) pairs ranked by entropy \(\mathcal{H}(\mathcal{O}_{s, a})\). This subset is labelled by a crowd of trainers with diverse skill levels (some might be adversarial). We estimate each trainer's reliability based on their feedback and weight their input accordingly when aggregating their feedback. Finally, the aggregated results are merged with the underlying RL policy $\pi_R(a \mid s)$, allowing us to sample an action to execute in the environment. This approach improves the efficiency of human feedback while accounting for varying trainer reliability.}
\label{fig:active_rhlf}
\end{figure*}

Figure~\ref{fig:active_rhlf} illustrates our proposed method. To evaluate our approach, we simulate human feedback using a pre-trained oracle and assess performance across a set of diverse domains, including classic 2D environments (Taxi, PACMAN, Frozen Lake) and a blood-glucose monitoring and control task for Type 1 Diabetes using the clinically approved UVA/Padova simulator \cite{DallaMan2014}. Experimental results show that our method significantly improves learning speed and feedback efficiency, particularly in deterministic settings where uncertainty is easier to quantify and resolve.

\section{Related Work}
\subsection{Learning From Crowds}
Learning from crowds is a well-established research area that addresses the challenge of aggregating noisy, unreliable, or biased labels provided by multiple trainers, often with varying expertise. A naive approach treats all trainers as equally reliable. However, performance can be improved by adjusting for trainer reliability \cite{snow2008cheap}.

One of the earliest attempts to compute trainer reliability explicitly was a probabilistic model that jointly estimates the true labels and the error rates of each trainer via Expectation-Maximization (EM) \cite{dawid1979maximum}. This foundational model has since inspired a range of probabilistic and Bayesian methods \cite{whitehill2009whose,raykar2010learning}, which learn trainer-specific confusion matrices and incorporate priors over worker accuracy.

Beyond probabilistic modelling, recent work leverages deep learning to model crowd annotations at scale. \citet{rodrigues2018deep} proposed learning with crowds using Gaussian processes and variational inference, while more recent neural architectures integrate crowd modelling directly into end-to-end training \cite{guan2018who,chu2020}, enabling better generalization in image classification and NLP tasks. In the reinforcement learning domain, \citet{siththaranjan2024distributionalpreferencelearningunderstanding} address the challenge of collecting preferences from humans with diverse preferences by modelling prefernce distributions and identifying hidden contexts and improving adversarial robustness.

\subsection{Active Learning}
Active learning (AL) is a supervised learning framework that aims to reduce annotation cost by selecting only the most informative data points for labelling. This selection is typically guided by an acquisition function that quantifies a model's uncertainty or expected gain from receiving a label. The foundational idea is that the learner queries those instances for which its current predictions are least confident. This approach, known as \emph{uncertainty sampling}, was introduced by~\citet{lewis1994sequentialalgorithmtrainingtext}. Subsequent strategies have proposed entropy-based selection (using Shannon information), margin sampling (difference between top prediction confidences), and expected model change (querying points likely to cause the greatest update to the model)~\cite{settles2009active}. These techniques have proven effective across a wide range of domains, including natural language processing, computer vision, and structured prediction.

While traditionally applied to classification tasks, active learning has recently seen broader use in more complex learning settings. Notably, several recent works have formulated active learning itself as a sequential decision-making process using reinforcement learning to learn acquisition policies~\cite{fang2017learningactivelearndeep,ebert2012}. Other works in the RLHF domain adopt reward model training as the target task, where feedback is derived from human preference annotations and refined through active sampling~\cite{ouyang2022traininglanguagemodelsfollow}.

In our approach, we use active learning only at the feedback selection stage: after each episode, the agent identifies the most uncertain state-action pairs and queries for feedback on only those. This selective feedback helps improve sample efficiency and avoids unnecessary burden on human trainers.

\subsection{Reinforcement Learning with Human Feedback}
Rather than relying solely on environment-provided rewards, Reinforcement learning with human feedback (RLHF) agents are guided by human evaluators through signals such as scalar feedback, preferences, or demonstrations. Early work in this area includes TAMER~\cite{Knox2012}, which proposed learning from real-time human-provided scalar feedback. Deep TAMER~\cite{warnell2018deep} extended this framework to high-dimensional continuous domains by using deep networks to model human responses. DQN-TAMER~\cite{arakawa2018dqn} further integrated TAMER-style feedback with standard deep RL objectives. Other methods such as Advise~\cite{Griffith2013} introduced the idea of policy shaping, directly modifying the agent’s action probabilities based on binary human input.
An influential shift came with preference-based methods~\cite{christiano2017deep}, where agents learn a reward model from human comparisons between trajectory segments, rather than receiving direct rewards or feedback. This line of work is exemplified by PEBBLE~\cite{lee2021pebble}, which first trains agents to explore a wide range of states and then elicits human preferences to learn a reward function. 
Recent work such as DUO~\cite{feng2025duo} advances this idea by incorporating active query selection into RLHF. DUO selects trajectory pairs based on informativeness, diversity, and proximity to the agent’s current policy (on-policy sampling), thereby improving the sample efficiency of reward learning. Similarly \citet{chhan2025crowdprefrlpreferencebasedrewardlearning} use methods derived from
unsupervised ensemble learning to effectively aggregate user preference
feedback across diverse crowds to learn RL policies. However, these approaches typically assume that no environment reward is available and thus focus entirely on constructing an accurate reward model from human input. 

By contrast, our approach complements the environment reward with targeted human feedback. We assume the agent already receives a reward signal from the environment, and our goal is to augment this with feedback on only the most uncertain state-action pairs. Unlike most RLHF approaches that model uncertainty at the level of the reward function, our method uses entropy derived from the value function to guide query selection.

\section{Preliminary}
\subsection{Reinforcement Learning}
\Ac{RL} is a framework for any learning process that involves sequentially interacting with an environment to achieve a certain objective~\cite{Sutton1998}. The learner is called the \textit{agent}. It performs an action $a_t$ in the \textit{environment}, observes its consequences $s_{t+1}$, and receives a reward $r_{t+1}$ (or a cost) signal -- a numerical assessment of the current situation as shown in the standard RL framework in Fig.~\ref{fig:active_rhlf}.

The agent basically learns a mapping from the state to the action that maximises the total amount of reward it receives over the long run. The mapping is called \textit{policy} denoted as $\pi_t\left(a \mid s\right)$ that indicates the probability of $a_t = a$ when the state is $s_t = s$.

\subsection{Advise Algorithm}
In order to incorporate feedback into the RL algorithm, we build upon the \textit{Advise} algorithm~\cite{Griffith2013} and thus provide a brief description of the approach. 
\textit{Advise} assumes binary feedback from a trainer that returns either `right' or `wrong' for a particular agent's choice of action. 
The feedback is accumulated for each state-action pair separately, and it is used to derive a trainer's policy, denoted by $\pi_{F}\left(a|s\right)$, which is then used to modify the agent's policy. 
Additionally, $C$ is \textit{consistency level}, defined as the probability that the trainer gives the right (consistent) feedback, and assuming a binomial distribution, the authors propose the following trainer policy.
\begin{equation} \label{eq:advise_t2}
\pi_{F}\left(a \mid s\right) \propto C^{\Delta(s,a)} (1-C)^{-\Delta(s,a)}
\end{equation} 
where $\Delta(s,a)$ is the difference between the number of positive and negative feedback from the trainer. 

The policy of the trainer is combined with $\pi_{R}\left(s,a\right)$ (policy from the underlying RL algorithm) by multiplying them together so that the final policy becomes as
\begin{equation} \label{eq:advise_shaping}
\pi \left(a \mid s\right) \propto \pi_{F}\left(a \mid s\right) \cdot \pi_{R}\left(a \mid s\right).
\end{equation}
This formulation enables direct shaping of the agent's behaviour through human feedback, rather than modifying the reward function. Such an approach is often considered more effective, as it allows the feedback to provide immediate influence on the policy itself.

\section{Method}
While \textit{Advise} provides a mechanism for incorporating human feedback, it exhibits three key limitations. First, it requires prior knowledge of the trainer's consistency level ($C$), which may be unknown or difficult to estimate in many practical applications. Second, the algorithm assumes a single trainer, limiting its scalability in settings involving multiple human supervisors or crowd-sourced feedback. Third, \textit{Advise} typically requires a large amount of feedback before producing noticeable improvements. While this is not necessarily a flaw of the algorithm itself, it presents a significant barrier in real-world scenarios where human feedback is costly and limited.

In this section, we propose a method to address all of the limitations above. Our approach supports feedback from multiple trainers and estimates each trainer's reliability (i.e., consistency level) in an online manner as feedback is collected (feedback from crowds). Additionally, the algorithm actively queries for feedback in situations where it is expected to be most informative (entropy-based active feedback). By estimating trainer reliability on the fly, the system leverages high-quality feedback while mitigating the impact of unreliable or adversarial trainers, making it robust to feedback of varying quality. Furthermore, by strategically selecting when to request feedback, the algorithm enhances the efficiency of human supervision, reducing the overall burden on the trainers.

\subsection{Feedback from Crowds}
Learning from Crowds is a well-established approach in supervised learning where labels are collected from a group of non-experts, reducing the burden of acquiring expert annotations. Typically, a probabilistic model iteratively estimates both the true labels and the reliability of each trainer to handle low-quality (noisy) annotators. We extend this framework to the reinforcement learning (RL) setting with introducing a \ac{VI} approach. This method iteratively estimates the distributions over both the optimal actions for given states and the reliability of individual trainers. The use of VI provides two key benefits: it allows the integration of prior knowledge, thereby improving the sample efficiency of the inference process, and it models uncertainty through parameter distributions, enhancing robustness in scenarios with sparse feedback.

Our VI-based approach infers the posterior distributions of trainer reliability (consistency level, $C_{l}$) and the optimality of state-action pairs ($\mathcal{O}{s,a}$) by maximizing the likelihood of human feedback observations. These observations are represented by $h^+_{l,s,a}$ and $h^-_{l,s,a}$, denoting the counts of positive and negative feedback, respectively, given by trainer $l$ for state $s$ and action $a$. The remainder of this section provides a high-level overview of our method. We refer the reader to Appendix A for additional details.

First, we define the posteriors using two variational factors $q(\mathbf{O}_s)$ and $q(C_l)$, as follows:
\begin{equation} \label{eq:rlhf_vi_likelihood_fact}
P\left(\mathbf{O}, \mathbf{C} | \mathbf{h}^{+}, \mathbf{h}^{-}\right) = \left(\prod_{s}q(\mathbf{O}_{s})\right) \left(\prod_{l} q(C_l) \right),
\end{equation}
where  $\mathbf{O}, \mathbf{C}, \mathbf{h}^+$ and $\mathbf{h}^-$ denote collections of  $\mathcal{O}_{s,a}, C_l, h^+_{l,s,a}$ and $h^-_{l,s,a}$, respectively for all trainers, states and actions. $\mathbf{O}_s$ represents the set of $\mathcal{O}_{s,a}$ values for all actions in a given state $s$.

The \ac{VI} approach updates the estimates of $q(\mathbf{O}_s)$ and $q(C_l)$ iteratively by maximising the evidence lower bound (ELBO) until convergence, after which we employ the \textit{posterior sampling}~\cite{Strens2000, thompson1933likelihood} by treating the estimated posterior $q(\mathbf{O}_s)$ as the policy:
 \begin{equation} 
    \label{eq:posterior_sampling_vi}
    \pi \left(a|s\right) = q\left(\mathbf{O}_s=\mathbf{e}_a\right),
\end{equation}
where $\mathbf{e}_a$ is a one-hot encoding vector. We take the underlying \ac{RL} policy as the prior of $q(\mathbf{O}_s)$. Consequently, it is naturally incorporated into the the posterior and the final policy is based on both interactions with environment and human feedback.

\subsection{Entropy-based Active Feedback}
Our method builds on an entropy-based measure of the posterior distribution over optimality variables for state-action pairs. Specifically, we compute the entropy of these variables along the experienced trajectory and prioritise feedback requests for the action pairs showing high entropy.

The posterior of the optimality variable $\mathcal{O}_{s,a}$ is conditioned on two types of observations: human feedback and interactions with the environment. Formally, the posterior is denoted as $P\left(\mathcal{O}_{s,a} | \mathbf{h}^+, \mathbf{h}^-, \tau\right)$, where the $\mathbf{h}^+$ and $\mathbf{h}^-$ represent positive and negative feedback, respectively and $\tau$ is a set of trajectories generated from interacting with the environment. Each trajectory $\tau$ consists of a sequence of state $s_t$, action $a_t$ and reward $r_t$ at each time step $t$ e.g. $\tau=\{s_0, a_0, r_0, s_1, a_1, r_1, \dots\}$.

Assuming conditional independence between human feedback $\mathbf{h}^+, \mathbf{h}^-$ and the trajectories $\tau$ given the optimality variable, the posterior can be factorised as follows (a detailed derivation is provided in Appendix B):
\begin{equation} \label{eq:posterior_opt_all}
\begin{split}    
    P\left(\mathcal{O}_{s,a} | \mathbf{h}^+, \mathbf{h}^-, \tau\right) \propto \qquad \quad &\\
    P\left(\mathcal{O}_{s,a} | \mathbf{h}^+, \mathbf{h}^-\right)& P\left(\mathcal{O}_{s,a} | \tau\right) / P\left(\mathcal{O}_{s,a} \right). 
\end{split}
\end{equation}
Here $P\left(\mathcal{O}_{s,a} \right)$ denotes the prior over the optimality. We assume a uniform prior, such that $P\left(\mathcal{O}_{s,a} = 1\right)=\frac{1}{N_a}$, where $N_a$ is a number of actions in a state.
The two posterior components -- one conditioned on human feedback and the other on interactions with the environment -- can be computed independently. These components are then combined multiplicatively, adjusted by the inverse of the prior, to yield the joint posterior. This structure facilitates scalable inference and enables integration of heterogeneous information sources.

In the following sections, we describe the procedures used to derive each of these posterior terms.

\subsubsection*{The posterior given the human feedback $P\left(\mathcal{O}_{s,a} | \mathbf{H}\right)$}
We build upon the results of \textit{Advise}~\cite{Griffith2013}, using its formulation to derive the posterior given human feedback, as expressed in the following equation:
\begin{equation} \label{eq:advise_t2_post_h}
P\left(\mathcal{O}_{s,a} | \mathbf{H}\right) \propto C^{\Delta(s,a)} (1-C)^{-\Delta(s,j)},
\end{equation} 
where $\Delta_{s,a}$ represents the difference between the number of positive and negative feedback on state $s$ and action $a$.

\subsubsection*{The posterior given the trajectories $P\left(\mathcal{O}_{s,a} | \tau\right)$}
We consider a baseline \ac{RL} algorithm that estimates the state-action value function $Q\left(s,a\right)$ from the trajectories. In this context, an action is considered optimal if it yields the highest value amongst all available actions in a given state. Therefore, we define the posterior distribution of the optimality variable as: 
\begin{equation}
    P\left(\mathcal{O}_{s,a}|\tau \right) = P\left(Q\left(s,a\right) > Q\left(s,a'\right) \text{ for all } a'\neq a\right).
\end{equation}
To make this computation tractable, we assume conditional independence of $Q\left(s,a'\right) \text{ for all } a' \neq a$ given $Q\left(s,a\right)$, and factorise each inequality. While this assumption is strong, it is reasonable if the actions are sufficiently distinct from other. This factorisation simplifies the posterior computation, as it allows us to express it as a product of independent terms. 
\begin{equation}
\begin{split} \label{eq:post_tau_2}
    P\left(\mathcal{O}_{s,a}|\tau \right)
    &= \int P\left(Q\left(s,a\right)=X\right) \cdot \\
    &\ \ \ \ \ \ \ \ \ \ P\left(X > Q\left(s,a' \right) \text{ for all } a'\neq a \mid X \right)\ dX\\
    &= \mathbb{E}_{X\sim P(Q(s,a))} \left[\prod_{a' \neq a} P\left(X > Q\left(s,a' \right)\right) \right] \\
    &= \mathbb{E}_{X\sim P(Q(s,a))} \left[\prod_{a' \neq a} F_{s,a'}(X) \right].
\end{split}
\end{equation}
Here, $F_{s,a'}\left(X\right)$ denotes the cumulative distribution function (CDF) of $P\left(Q\left(s,a'\right)\right)$.

Next, we assume that $P\left(Q\left(s,a\right)\right)$ follows Gaussian distribution with the mean $\hat{Q}\left(s,a\right)$ and the standard deviation $\sigma_{base}/\sqrt{N_{s,a}}$, where $\hat{Q}\left(s,a\right)$ is the estimated value function, $\sigma_{base}$ is a baseline standard deviation (hyper-parameter) and $N_{s,a}$ is a number of visits to the state-action pair.

We compute the expectation in Eq.~\ref{eq:post_tau_2} using Monte Carlo estimation. Specifically, we apply the inverse transform method with stratified sampling. First, we construct a sequence of $M$ evenly spaced probability values:
\begin{equation}
    p_i=\frac{1}{2M}+\frac{i}{M} \text{ for }i=0,1,2,\dots,M-1.
\end{equation}
Then, we obtain the samples of $Q(s,a)$ by applying the inverse CDF:
\begin{equation}
    x_i=F^{-1}_{s,a}(p_i) \text{ for } i=0,1,2,\dots,M-1.
\end{equation}
Finally, we approximate the posterior using:
\begin{equation} \label{eq:post_opt_tau_approx}
     P\left(\mathcal{O}_{s,a}|\tau \right) \simeq \frac{1}{M} \sum_{i=0}^{M-1} \prod_{a' \neq a} F_{s,a'}(x_i).
\end{equation}
Since we assume a Gaussian distribution for $Q(s,a)$, both $F^{-1}_{s,a}(p_i)$ and $F_{s,a'}(x_i)$ can be computed analytically. In addition, the Monte Carlo estimation process lends itself well to parallel (vectorised) computation, which improves efficiency and reduces runtime.

\subsubsection*{One-vs-All Entropy}
We now have the posterior distribution $P\left(\mathcal{O}_{s,a} | \mathbf{h}^+, \mathbf{h}^-, \tau\right)$ for each state-action pair, and we are ready to compute its entropy. However, the conventional Shannon entropy is not well-suited in this context. The variable $\mathcal{O}_{s,a}$ is binary where $\mathcal{O}_{s,a}=1$ indicates the action $a$ is optimal at state $s$, and $\mathcal{O}_{s,a}=0$ implies that the other action(s) is optimal. Consequently, we require an entropy measure $\mathcal{H}$ that satisfies the following criteria (Requirement IV is desirable but not essential):
\begin{enumerate}[label=\Roman*.]
    \item $\mathcal{H}\left(\mathcal{O}_{s,a}\right) \rightarrow 0$ when $P\left( \mathcal{O}_{s,a} = 1\right) \rightarrow 0$.
    \item $\mathcal{H}\left(\mathcal{O}_{s,a}\right) \rightarrow 0$ when $P\left( \mathcal{O}_{s,a} = 1\right) \rightarrow 1$.
    \item $\mathcal{H}\left(\mathcal{O}_{s,a}\right)$ reaches its max. when $P\left( \mathcal{O}_{s,a} = 1\right) = 1/N_a$.
    \item $\mathcal{H}\left(\mathcal{O}_{s,a}\right)$ reduces to the Shannon entropy when $N_a=2$.
\end{enumerate}

To meet these requirements, we propose the One-vs-All (OvA) entropy. The OvA entropy is defined for a binary variable ($\mathcal{O}_{s,a}$) with Bernoulli distribution (parameter $p$) and it is computed in two steps.
\begin{enumerate}
    \item Re-normalise Bernoulli parameter $p$ using: $p'=\frac{p}{p + (1-p)/(N_a -1)}$.
    \item Compute Shannon entropy for Bernoulli distribution with the re-nomalised parameter $p'$.
\end{enumerate}
The resulting OvA entropy $\mathcal{H}_{ova}$ is given by:
\begin{equation}
    \mathcal{H}_{ova}\left(\mathcal{O}_{s,a}\right)=-p' \log_2\left(p'\right) - (1-p')\log_2\left(1-p'\right),
\end{equation}
where $p'=\frac{p}{p + (1-p)/(N_a -1)}$ and $p=P\left(\mathcal{O}_{s,a}=1 | \mathbf{h}^+, \mathbf{h}^-, \tau\right)$. The re-normalisation scales down the non-optimal probability $1-p$ by the number of other actions $N_a-1$. As the result, the OvA entropy $\mathcal{H}_{ova}$ gives the maximum value when $p=1/N_a$. This form of entropy is related in nature to the family of asymmetric entropies \cite{GUERMAZI2018373, santos2009cost}.

\subsubsection*{Summary}
The entropy-based active feedback method proceeds as follows: first, the agent collects trajectories (state-action pairs) through interaction with the environment. After completing an episode (or after a fixed number of episodes), the method computes the posteriors of the optimality variable for all state-action pairs encountered in the trajectories, using Eq.~\ref{eq:posterior_opt_all}, Eq.~\ref{eq:advise_t2_post_h}, and Eq.~\ref{eq:post_opt_tau_approx}. Next, the OvA entropy is calculated for each state-action pair. Finally, the method identifies the state-action pairs with the top highest OvA entropies and requests human feedback for these pairs.

This method strategically prioritises querying human feedback for state-action pairs exhibiting high uncertainty with respect to their optimality. As a result, the system maximises the information gain from human feedback by targeting ambiguous or uncertain decisions. Moreover, by limiting the scope to state-action pairs encountered within actual agent trajectories, the approach concentrates learning on decision points that are  relevant and likely to recur during future interactions with the environment.

\section{Experiments}
We conduct experiments across a diverse set of environments that operate on a gridworld domain; these include our own PACMAN-inspired gridworld as well as Taxi and FrozenLake from the Farama Gymnasium suite \cite{towers2024gymnasiumstandardinterfacereinforcement}. We then test our method on a real-world task; blood glucose control for Type 1 Diabetes using the clinically validated UVA/Padova simulator. Across all environments, we compare our entropy-based active querying and crowd feedback aggregation approach against (1) the baseline approach (no feedback) and (2) random active learning (feedback randomly sampled from trajectories). We train an oracle policy using standard reinforcement learning on the target environment as a proxy for an expert human. We modify $C_l = [0.9,0.8,0.6,0.3] $ across 4 trainers to simulate human trainers of varying quality. 

\subsection{Gridworld Domains}
We first implemented a simplified abstraction of the classic PACMAN game, using a small (5x5) grid and a reduced state space consisting of two pellets and a single ghost. This controlled setup allows us to test agent learning in environments that require sequential planning, without the full complexity of the original game. 

\begin{figure}[t]
\centering
\includegraphics[width=\columnwidth]{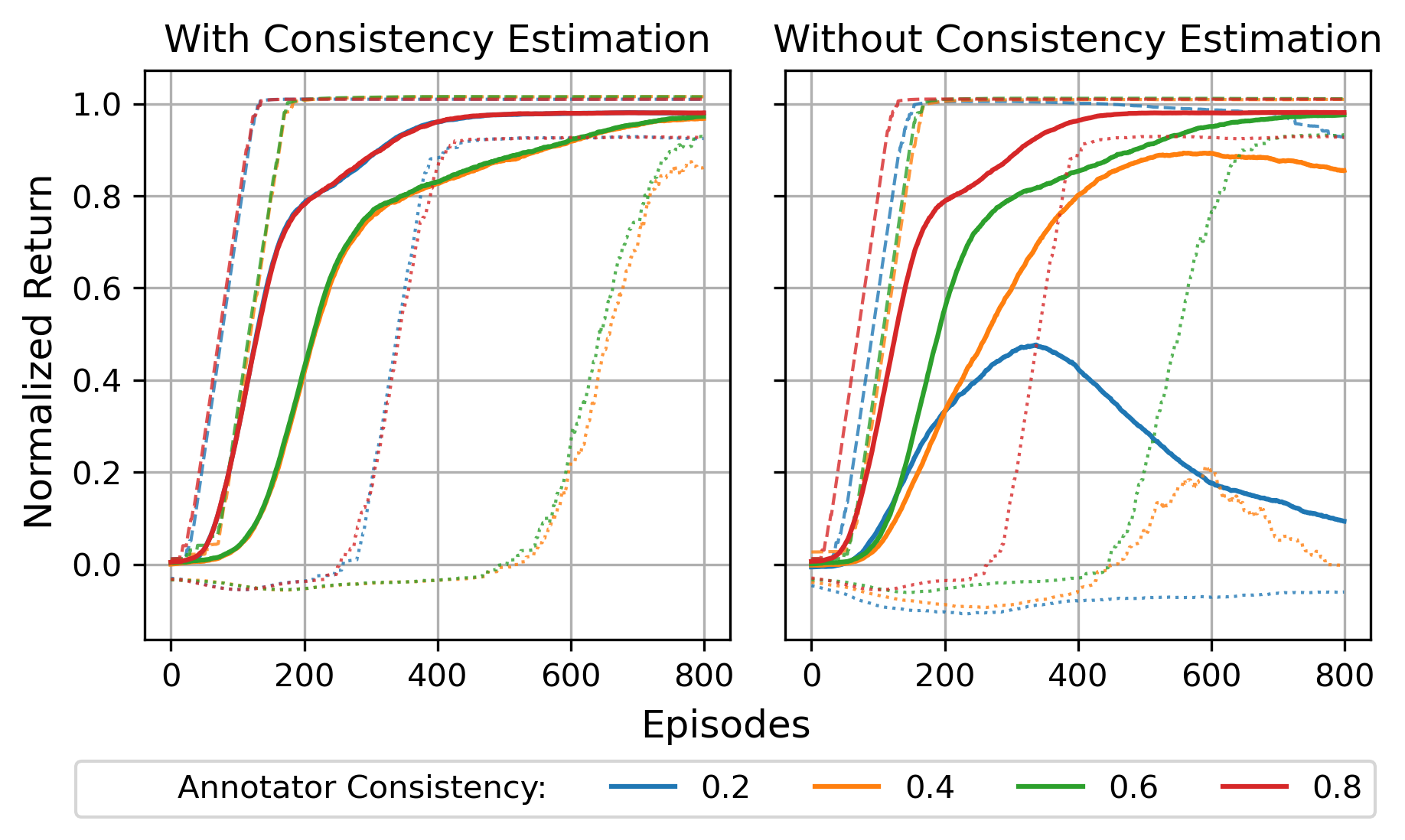} 
\caption{Evaluation results with and without the consistency level estimation. The agent without the estimation assumes the consistency level is 0.8 fixed, while varying the true value over 0.8, 0.6, 0.4 and 0.2. Dashed and dotted lines correspond to the 95\% upper and 5\% lower confidence bounds, respectively.}
\label{fig:pacman_cfix_vs_cest}
\end{figure}

We validated our \textbf{consistency level estimation} method in this PACMAN environment.
After confirming the effectiveness, we expanded our evaluation to additional gridworld domains with varying dynamics and challenges:
\begin{itemize}
    \item \textbf{Taxi environment}: complementary evaluation featuring random starting positions and destinations and a larger map. Default parameters from the gymnasium library were used without modification
    \item \textbf{Customised Frozen Lake environments}: designed to investigate the relationship between task difficulty and performance of our active learning method.
\end{itemize}

For each environment, we run 50 trials of 1000 episodes per agent and take the average return to account for training noise. Our learning rate is set to $\alpha = 0.05$ and discount factor $\gamma = 0.9$. Detailed environment descriptions are provided in the technical appendix. Our code and experiments will be made publicly available in the camera-ready version.

\begin{figure}[t]
\centering
\includegraphics[width=\columnwidth]{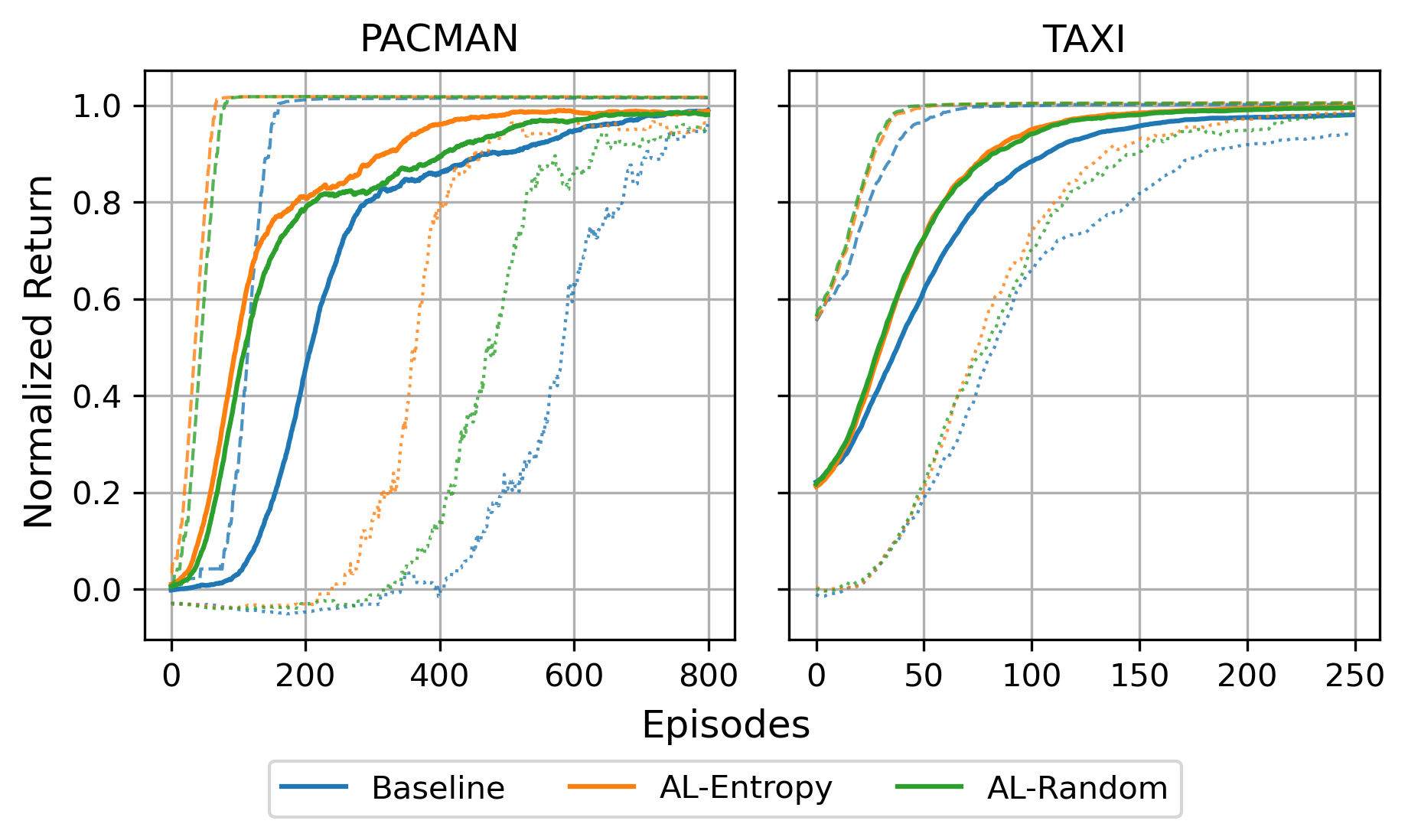} 
\caption{Evaluation results for PACMAN and Taxi gridworld environments. Agents are trained with standard Q-Learning (Baseline) and active learning with random sampling (AL-Random) and entropy-based sampling (AL-Entropy). Dashed and dotted lines correspond to the 95\% upper and 5\% lower confidence bounds, respectively.}
\label{fig:pacman_taxi}
\end{figure}

\begin{figure}[t]
\centering
\includegraphics[width=\columnwidth]{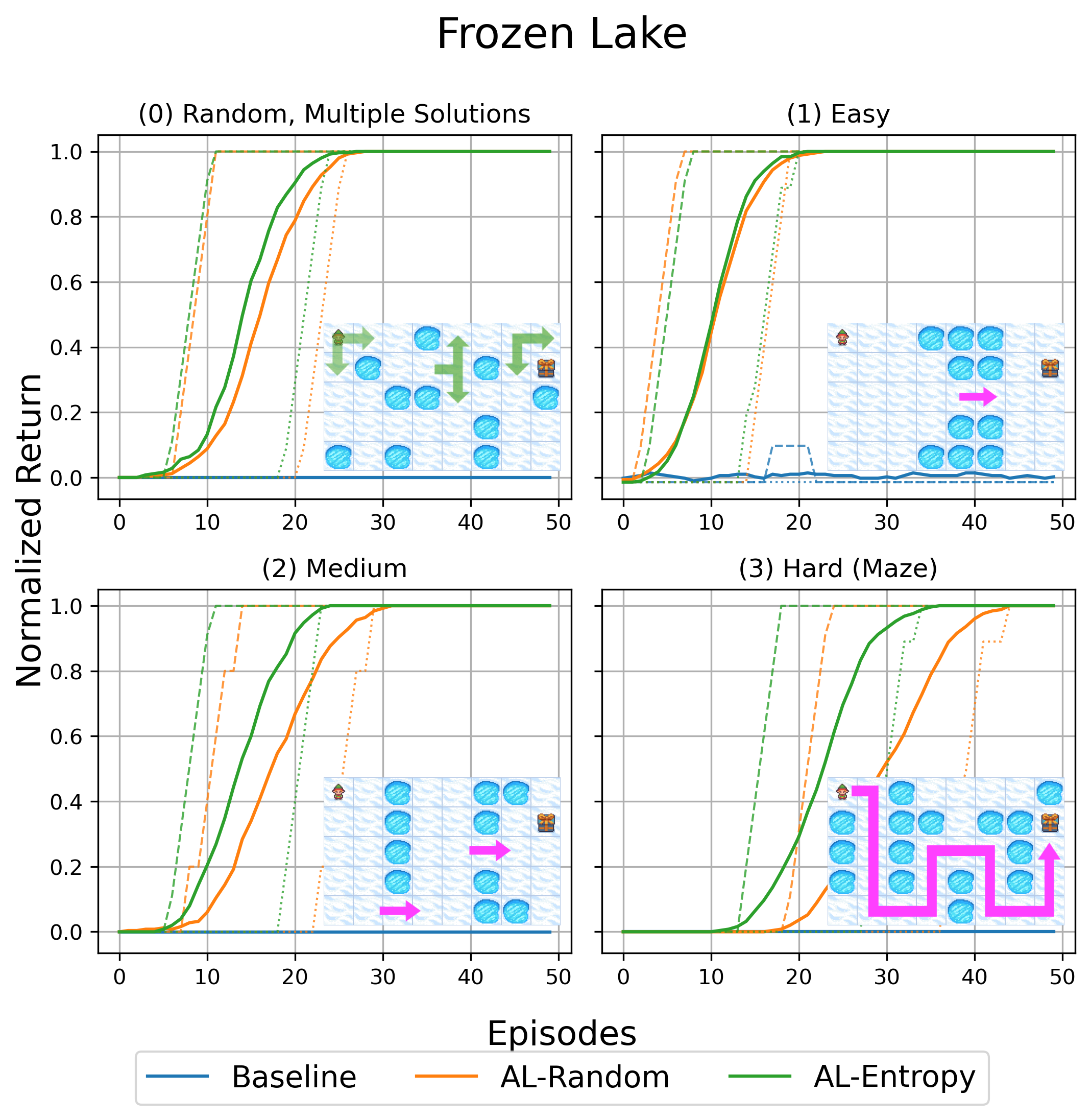} 
\caption{Evaluation results for the Frozen Lake gridworld environment. Agents are trained with standard Q-Learning (Baseline) and active learning with random sampling (AL-Random) and entropy-based sampling (AL-Entropy). Dashed and dotted lines correspond to the 95\% upper and 5\% lower confidence bounds, respectively.}
\label{fig:frozen_lake}
\end{figure}

\paragraph{Results}

We validated our \textbf{consistency level estimation} method in the PACMAN environment by comparing agents with and without this feature without using our active feedback. In the non-estimating condition, we fix the consistency level at 0.8, while varying the true value across 0.8, 0.6, 0.4, and 0.2.
Figure~\ref{fig:pacman_cfix_vs_cest} shows that agents without estimation perform poorly when the assumed and true consistency differ. Agents with the ability to estimate consistency perform well across all values. The best overall performance occurs when the true level is 0.8 (which is matched with the assumed level) and estimation is disabled.

These results demonstrate that consistency level estimation enhances robustness when the agent receives feedback from trainers with unknown  reliability. However, if the trainer's consistency level is known, directly using the known value yields better performance than estimation.

We now extend our evaluation to our full suite of environments incorporating both consistency level estimation and active feedback mechanisms.
Figure~\ref{fig:pacman_taxi} shows learning curves for PACMAN and Taxi. In PACMAN, our entropy-based active learning method (AL-Entropy) outperforms both the baseline and random sampling (AL-Random). In contrast, results on Taxi are less conclusive. We hypothesise that this discrepancy is due to the structural differences between the environments: Taxi presents a relatively unconstrained space with many viable paths to success, while PACMAN imposes stricter conditions for winning; particularly due to board sections where the goal (a pellet) is difficult to reach without navigating around a ghost.

To explore this hypothesis, we test four variants of the Frozen Lake environment: one with a randomly generated map offering multiple solution paths, and three increasingly constrained maps where successful navigation depends on passing through one or more mandatory "gates".

Figure~\ref{fig:frozen_lake} illustrates the learning curves for entropy-based sampling vs. random sampling in these environments. Both methods perform similarly in the open, unconstrained setting. There is no discernable improvement from random sampling in the easy environment (1). However, as the environment becomes more constrained, entropy-based sampling increasingly outperforms random sampling. Table~\ref{tab:gridworld_results} shows the difference in area under the curve (AUC) for each of our experiments. A larger AUC implies faster learning.

\begin{table}[t]
\setlength{\tabcolsep}{1mm}
\centering
\begin{tabular}{lcccc}
\toprule
\textbf{Environment} & \textbf{Baseline} & \textbf{AL-Random} & \textbf{AL-Entropy} \\
\midrule
PACMAN              & 546.91  & 632.02  & \textbf{660.77}  \\
\midrule
Taxi                & 201.33  & 209.19  & \textbf{209.75}  \\
\midrule
Frozen Lake (0) & 0.00    & 32.88  & \textbf{34.76} \\
Frozen Lake (1) & 0.24    & 38.30  & \textbf{38.59} \\
Frozen Lake (2) & -0.01   & 31.38  & \textbf{35.14} \\
Frozen Lake (3) & 0.00    & 19.12  & \textbf{26.40} \\
\bottomrule
\end{tabular}%
\caption{Area Under the Curve (AUC) comparison across gridworld environments and methods. Higher values indicate better performance.}
\label{tab:gridworld_results}

\end{table}

\subsection{Type 1 Diabetes BGL Control}
We evaluate our approach in a clinically relevant healthcare application: the control of \ac{BGL} in individuals with type 1 diabetes. The objective of this task is to maintain \ac{BGL} within a healthy range by administering insulin at the appropriate dosage and timing. The agent monitors \ac{BGL} through \ac{CGM} and receives information about carbohydrate intake (i.e., meals). Based on these observations, it determines both the timing and amount of insulin to be delivered.

For this evaluation, we employ the UVA/Padova type 1 diabetes simulator~\cite{DallaMan2014}, the first computational model approved by the U.S. Food and Drug Administration (FDA) as a substitute for preclinical trials of certain insulin treatments. We use an open-source implementation of this simulator~\cite{Xie2018}, which includes physiological profiles of 30 distinct virtual individuals. The simulator models the physiological response of \ac{BGL} to insulin administration, allowing us to systematically assess the agent's performance in a realistic and high-fidelity simulation environment.

\paragraph{Results}
We evaluate three agent models for this task: standard Q-learning (Baseline), Q-learning with feedback with uniformly random sampling (AL-Random), and Q-learning with entropy-based active feedback (AL-Entropy). We test these models on three virtual individuals provided by the UVA/Padova simulator. Each model is trained on 2,000 episodes with five random seeds, and performance is measured by the percentage of time that \ac{BGL} remain within the target range of 70–180 mg/dL during the training period. 

The results, presented in Table~\ref{tab:diabetes_results}, show that the entropy-based active feedback model consistently outperforms the other baselines across subjects, confirming the benefits of the approach on a challenging safety-critical task.

\begin{table}
    \setlength{\tabcolsep}{1mm}
    \centering
    
    \begin{tabular}{cccc}
        \toprule
        \textbf{Individual} & \textbf{Baseline} & \textbf{AL-Random} & \textbf{AL-Entropy} \\
        \midrule
        child003      & $36.7 \pm 1.4$ & $49.5 \pm 1.6$ & $\bf{54.4} \pm 1.1$\\
        adolescent003 & $45.7 \pm 3.6$ & $69.6 \pm 1.9$ & $\bf{75.6} \pm 0.9$\\
        adult003      & $60.6 \pm 4.3$ & $80.4 \pm 3.7$ & $\bf{88.6} \pm 1.7$\\
         \bottomrule
    \end{tabular}
    \caption{\% of time in the target BGL range (70-180 mg/dL) across entire training episodes (2000) with the standard deviation. The best results amongst the three models are highlighted in bold face.}
    \label{tab:diabetes_results}
\end{table}

\section{Discussion}
Our gridworld experiments enabled us to test environments with different dynamics. We propose that in environments with many good solutions, the uncertainty in any given state might be spread across multiple acceptable actions; the entropy for each action might be relatively high, but not because one action is definitively better than another. In such a scenario, randomly sampling might be just as effective, or even more efficient than calculating entropy. Entropy-based active learning surpasses random sampling when the problem becomes more constrained; our superior performance on the blood glucose control task aligns with our conclusion that active learning is well-suited to these highly-constrained tasks.

\paragraph{Limitations} The additional steps in our pipeline come at a cost; whether it is the computational effort required to process and downsample trajectories, or the time required by a (human or synthetic) teacher to annotate these trajectories, we must consider the trade-off vs. the cost of additional training time. Additionally, we assume that we have readily available expert and non-expert teachers.

Real humans could potentially lead to injecting biases into our models; any real-world application would require testing against adversarial teachers. There is also a known trade-off between the sample efficiency of active learning and the robustness of the resulting policy. While our method improves learning speed, we have yet to establish a methodology for evaluating robustness in this context.

Finally, while we have validated our method on a real-world control task, its implementation relies on tabular Q-learning; a simple and limited learning algorithm that does not scale well to more complex or high-dimensional environments.
\paragraph{Future Work}
To assess the broader applicability of our approach, future work will explore its integration with function approximation methods such as Deep Q-Networks (DQNs). 

Beyond scalability, agent stability warrants further investigation alongside consideration of recent methods to enhance robustness \cite{yamagata2024saferobustreinforcementlearning}. 

Finally, we aim to implement human data collection to better understand the relationship between annotation effort and performance improvement.

\bibliography{aaai2026}
\clearpage

\acrodef{RL}{reinforcement learning}
\acrodef{VI}{variational inference}
\acrodef{MI}{mutual information}
\acrodef{BSC}{binary symmetric channel}
\acrodef{BGL}{blood glucose level}
\acrodef{CGM}{continuous glucose monitor}
\acrodef{BBController}{Basal-Bolus controller}

\newif\ifreproStandalone
\reproStandalonefalse
\def\isChecklistMainFile{}  

\newpage

\makeatletter
\@ifundefined{isChecklistMainFile}{
  \newif\ifreproStandalone
  \reproStandalonetrue
}{
  \newif\ifreproStandalone
  \reproStandalonefalse
}
\makeatother

\ifreproStandalone

\documentclass[letterpaper]{article}
\usepackage[hyphens]{url}  
\usepackage{graphicx} 
\urlstyle{rm} 
\def\UrlFont{\rm}  
\usepackage{natbib}  
\usepackage{caption} 

\usepackage[submission]{aaai2026}
\setlength{\pdfpagewidth}{8.5in}
\setlength{\pdfpageheight}{11in}
\usepackage{times}
\usepackage{helvet}
\usepackage{courier}
\usepackage{colortbl} 
\usepackage[table]{xcolor}
\usepackage{graphicx}
\usepackage{fancyvrb}
\frenchspacing

\usepackage{xr} 

\usepackage{algorithm}
\usepackage{algorithmic}

\usepackage{amsmath, amssymb}
\usepackage[nolist]{acronym}
\usepackage{booktabs}
\usepackage{enumitem}
\usepackage{colortbl} 

\usepackage{subcaption}

\usepackage{array}

\usepackage{float}



\acrodef{RL}{reinforcement learning}
\acrodef{VI}{variational inference}
\acrodef{MI}{mutual information}
\acrodef{BSC}{binary symmetric channel}
\acrodef{BGL}{blood glucose level}
\acrodef{CGM}{continuous glucose monitor}
\acrodef{BBController}{Basal-Bolus controller}

\externaldocument{main}

\begin{document}
\fi

\setlength{\leftmargini}{20pt}
\makeatletter\def\@listi{\leftmargin\leftmargini \topsep .5em \parsep .5em \itemsep .5em}
\def\@listii{\leftmargin\leftmarginii \labelwidth\leftmarginii \advance\labelwidth-\labelsep \topsep .4em \parsep .4em \itemsep .4em}
\def\@listiii{\leftmargin\leftmarginiii \labelwidth\leftmarginiii \advance\labelwidth-\labelsep \topsep .4em \parsep .4em \itemsep .4em}\makeatother

\setcounter{secnumdepth}{0}
\renewcommand\thesubsection{\arabic{subsection}}
\renewcommand\labelenumi{\thesubsection.\arabic{enumi}}

\newcounter{checksubsection}
\newcounter{checkitem}[checksubsection]

\newcommand{\checksubsection}[1]{%
  \refstepcounter{checksubsection}%
  \paragraph{\arabic{checksubsection}. #1}%
  \setcounter{checkitem}{0}%
}

\newcommand{\checkitem}{%
  \refstepcounter{checkitem}%
  \item[\arabic{checksubsection}.\arabic{checkitem}.]%
}
\newcommand{\question}[2]{\normalcolor\checkitem #1 #2 \color{blue}}
\newcommand{\ifyespoints}[1]{\makebox[0pt][l]{\hspace{-15pt}\normalcolor #1}}

\setcounter{figure}{4}
\setcounter{equation}{12}
\setcounter{table}{2}

\onecolumn
\newpage
\appendix
\section{Appendix A: Variational Inference Approach}
\label{app:vi_algorithm}
We introduce our \ac{VI} algorithm for the consistency level estimation. The \ac{VI} algorithm estimates the distribution of the consistency level. Hence, it captures the uncertainty of the estimation, and we expect it to be more robust against potential estimation errors, especially when we do not receive enough feedback from the trainer. Also, \ac{VI} could incorporate prior information, and it potentially dramatically improves the estimation accuracy.

To clarify the assumptions underlying our formulation, we introduce a graphical model that represents all relevant random variables and their dependencies. This visualization provides a concise overview of the probabilistic relationships of our framework.
Before presenting the graphical model, we define the symbols used in the diagram for clarity and consistency.

\begin{figure}[h!]
	\centering
	\includegraphics[width=0.7\textwidth]{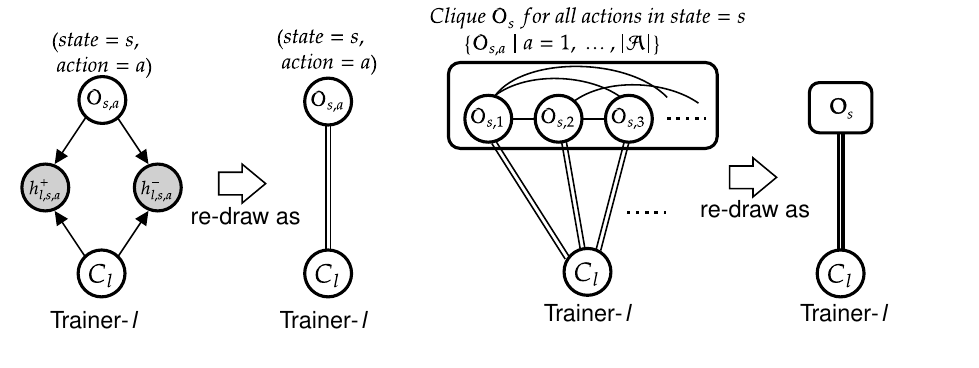}
	\caption{Symbols for graphical model for consistency level estimation with VI algorithm.}
	\label{fig:graph_symbol_vi}
\end{figure}

We assume that the number of positive and negative feedback for a given trainer and state-action pair ($h_{l,s,a}^+$ and $h_{l,s,a}^-$) are depending upon the optimality flag for the state-action pair ($\mathcal{O}_{s,a}$) and the trainer's consistency level ($C_l$), and their relationship can be drawn as the left-most figure in Fig.~\ref{fig:graph_symbol_vi}. We redraw it with the double line between $\mathcal{O}_{s,a}$ and $C_l$ as the second left figure in Fig.~\ref{fig:graph_symbol_vi}.
Next, we introduce a clique that has all optimality flags for a given state. The optimality flags in the clique are fully connected. Then, we rewrite the clique with the rounded box as shown in the rightmost figure in Fig.~\ref{fig:graph_symbol_vi}. Also, the connections between $C_l$ and the optimality flags are replaced with a bold double line, as shown in the figure. Note that the bold double line has all related observed variables -- the number of positive and negative feedback for all actions from a given trainer $l$ in a state $s$.

\begin{figure}[h!]
	\centering
	\includegraphics[width=0.7\textwidth]{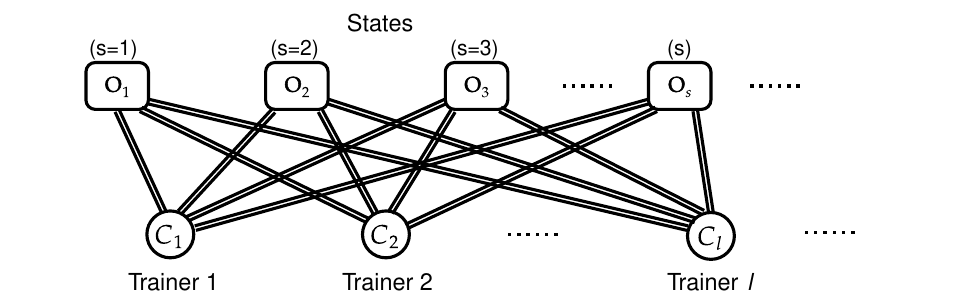}
	\caption{Graphical model for consistency level estimation with VI algorithm.}
	\label{fig:graph_type2_vi}
\end{figure}

With the above symbols, we can draw our graphical model for the random variables as Fig.~\ref{fig:graph_type2_vi}.

Based on the above graphical model, the complete likelihood function can be factorised as,
\begin{equation} \label{eq:rlhf_vi_likelihood}
P\left(\mathbf{h}^{+}, \mathbf{h}^{-}, \mathbf{O}, \mathbf{C} \right) = \left(\prod_{s,a}\prod_l P(h_{l,s,a}^+, h_{l,s,a}^- | \mathcal{O}_{s,a}, C_l)\right) \left(\prod_{s} P(\mathbf{O}_s) \right) \left(\prod_{l} P(C_l) \right),
\end{equation}
The hidden variable $\mathbf{O}$ is a matrix, and each element is an optimality flag $\mathcal{O}_{s,a}$, which is a boolean that is 1 when $a$ is the optimal action at state $s$, and 0 otherwise.
The conditional probability of $h_{l,s,a}^+$ and $h_{l,s,a}^+$ is the binomial probability as defined as follows:
\begin{equation} \label{eq:rlhf_binominal}
P\left( h_{l,s,a}^{+}, h_{l,s,a}^{-} | \mathcal{O}_{s,a} ; C_l)\right) = 
\begin{cases}
    \begin{pmatrix} h_{l,s,a}^{+} + h_{l,s,a}^{-} \\ h_{l,s,a}^{+} \end{pmatrix} C_l^{h_{l,s,a}^{+}} \left(1-C_l\right)^{h_{l,s,a}^{-}}
            & \mbox{if } \mathcal{O}_{s,a} = 1 \\
    \begin{pmatrix} h_{l,s,a}^{+} + h_{l,s,a}^{-} \\ h_{l,s,a}^{-} \end{pmatrix} \left(1-C_l\right)^{h_{l,s,a}^{+}} C_l^{h_{l,s,a}^{-}}
            & \mbox{if } \mathcal{O}_{s,a} = 0,
\end{cases}
\end{equation}
and the prior of $\mathbf{O}_s$ is set to the underlying \ac{RL} algorithm policy ($P(\mathbf{O}_s=\mathbf{e}_a)=\pi_R(a|s)$). $P(C_l)$ is the prior for the $l$-th trainer's consistency level. We employ the conjugate prior (the beta distribution) for $C_l$, and it is defined as,
\begin{equation} \label{eq:rlhf_vi_beta_prior}
P\left(C_l\right) = \frac{\Gamma(\alpha_l+\beta_l)}{\Gamma(\alpha_l)\Gamma(\beta_l)} C_l^{\alpha_l-1} (1-C_l)^{\beta_l-1},
\end{equation}
where $\Gamma$ is the Gamma function. We approximate the posterior of $\mathbf{O}$ and $\mathbf{C}$ with two variational factors as follows, and \ac{VI} performs approximate inference by updating each variational factor, $q(z)$, in turn, optimising the approximate posterior distribution until it converges.
\begin{equation} \label{eq:rlhf_vi_likelihood_fact}
P\left(\mathbf{O}, \mathbf{C} | \mathbf{h}^{+}, \mathbf{h}^{-}\right) = \left(\prod_{s}q(\mathbf{O}_{s})\right) \left(\prod_{l} q(C_l) \right),
\end{equation}

The variational factor $q(C_l)$ is updated by taking an expectation over the current estimate of the other variational factor $q(\mathbf{O}_s)$.
\begin{equation} \label{eq:rlhf_vi_update_q(c)}
\log q\left(C_l\right) = \left(h_l^r + \alpha_l - 1\right) \log C_l + \left(h_l^w + \beta_l - 1\right) \log (1-C_l) + \text{Constant},
\end{equation}
where the $h_l^r$ and $h_l^w$ are the expected number of right and wrong feedback, respectively. They are derived as,
\begin{equation}
\begin{split}
    \label{eq:rlhf_vi_hr_hw}
    h_l^r &= \sum_{s,a} q\left(\mathcal{O}_{s,a}=1\right) h_{l,s,a}^+ + q\left(\mathcal{O}_{s,a}=0\right) h_{l,s,a}^- \\
    h_l^w &= \sum_{s,a} q\left(\mathcal{O}_{s,a}=0\right) h_{l,s,a}^+ + q\left(\mathcal{O}_{s,a}=1 \right) h_{l,s,a}^-.
\end{split}
\end{equation}
Constant in Eq.~\ref{eq:rlhf_vi_update_q(c)} is the partition function for $\log q(C_l)$. As it is a log of the beta distribution, The partition function becomes $\log \left(\frac{\Gamma(h_l^r+h_l^w+\alpha_l+\beta_l)}{\Gamma(h_l^r+\alpha_l)\Gamma(h_l^w+\beta_l)}\right)$.
The other variational factor $q(\mathbf{O}_s)$ is updated as,
\begin{equation} 
\begin{split}
\label{eq:rlhf_vi_update_q(o)}
\log q\left(\mathbf{O}_s=\mathbf{e}_a\right) = 
    \sum_l \bigg\{ \delta_{l,s,a} \mathbb{E}\log C_l -
                   \delta_{l,s,a} \mathbb{E}\log (1-C_l) \bigg\}
                + \log \pi_R(a|s) 
                + \text{Constant}.
\end{split}
\end{equation}
Where the $\mathbb{E}\log C_l$ and $\mathbb{E}\log (1-C_l)$ are the expected log of the consistency level and one minus consistency level for the trainer $l$. They are derived as,
\begin{equation} 
\begin{split}
\label{eq:rlhf_vi_update_ElogC}
    \mathbb{E}\log C_l     &= \psi(h_l^r+\alpha_l) - \psi(h_l^r+h_l^w+\alpha_l+\beta_l) \\
    \mathbb{E}\log (1-C_l) &= \psi(h_l^w+\beta_l)  - \psi(h_l^r+h_l^w+\alpha_l+\beta_l),
\end{split}
\end{equation}
where $\psi$ is the digamma function. Constant in Eq.~\ref{eq:rlhf_vi_update_q(o)} is the partition function to make $\sum_a \log q(\mathbf{O}_s=\mathbf{e}_a) = 0$ for $\forall s$. 
Then, it repeats updating the variational factors until they converge. Once it is converged, we employ Eq.~\ref{eq:rlhf_vi_update_q(o)} to derive the posterior of $\mathbf{O}_s$ and use it as the policy to decide the following action.
The overall \ac{VI} algorithm is summarised in Algorithm~\ref{alg:rlhf_vi}.

\begin{algorithm}
\caption{Consistency Level Estimation with VI algorithm}
\label{alg:rlhf_vi}
\begin{algorithmic}[1]
\REQUIRE $i_{max}$, $\alpha$, $\beta$, $\pi_R(a|s)$, $\mathbf{h}^{+}$ and  $\mathbf{h}^{-}$ 
\STATE $\delta(l,s,a) \leftarrow h_{l,s,a}^{+} - h_{l,s,a}^{-}$ for $\forall l, \forall s, \forall a$
\STATE $i \leftarrow 1$
\STATE Initialise $q(\mathbf{O}_s=\mathbf{e}_a)\leftarrow\pi_R(a|s)$ for $\forall s, \forall a$
\WHILE{TRUE}
  \STATE Update $h_l^r$ and $h_l^w$ for $\forall l$ with Eq.~\ref{eq:rlhf_vi_hr_hw}.
  \STATE Update $\log q(C_l)$ for $\forall l$ with Eq.~\ref{eq:rlhf_vi_update_q(c)}.
  \STATE Update $\mathbb{E}\log C_l$ and $\mathbb{E}\log (1-C_l)$ for $\forall l$ with Eq.~\ref{eq:rlhf_vi_update_ElogC}.
  \STATE Update $\log q(\mathbf{O}_s=\mathbf{e}_a)$ for $\forall s, \forall a$ with Eq.~\ref{eq:rlhf_vi_update_q(o)}.
  \IF{$\text{converged}(\mathbb{E}\log C_l, \mathbb{E}\log (1-C_l), \forall l)$ or $i == i_{max}$}
    \STATE break
 \ENDIF
 \STATE $i \leftarrow i+1$
\ENDWHILE
\RETURN $\mathbb{E}\log C_l$ and $\mathbb{E}\log (1-C_l)$ for $\forall l$
\end{algorithmic}
\end{algorithm}

\section{Appendix B: Derivation of Eq.~\ref{eq:posterior_opt_all}}
This appendix provides a detailed derivation of the posterior factorisation of the optimality variable, as expressed in Eq.~\ref{eq:posterior_opt_all}. The derivation relies on the application of Bayes' rule in the first and third lines, as shown below. We also assume conditional independence between human feedback $\mathbf{h}^+, \mathbf{h}^-$ and the trajectories $\tau$, given the optimality variable $\mathcal{O}_{s,a}$.

\begin{equation}
\begin{split}
P\left(\mathcal{O}_{s,a} | \mathbf{h}^+, \mathbf{h}^-, \tau\right)  
    &= P\left(\mathbf{h}^+, \mathbf{h}^-, \tau \mid \mathcal{O}_{s,a} \right) \frac{P\left( \mathcal{O}_{s,a} \right)}{P\left( \mathbf{h}^+, \mathbf{h}^-, \tau \right)} \\
    &= P\left(\mathbf{h}^+, \mathbf{h}^- \mid \mathcal{O}_{s,a} \right) P\left( \tau \mid \mathcal{O}_{s,a} \right) \frac{P\left( \mathcal{O}_{s,a} \right)}{P\left( \mathbf{h}^+, \mathbf{h}^-, \tau \right)} \quad (\because \text{the conditional independence})\\
    &= \frac{P\left(\mathcal{O}_{s,a} | \mathbf{h}^+, \mathbf{h}^-\right) P\left( \mathbf{h}^+, \mathbf{h}^-\right)}{P\left(\mathcal{O}_{s,a}\right)} \cdot \frac{P\left(\mathcal{O}_{s,a} | \tau \right) P\left( \tau \right)}{P\left(\mathcal{O}_{s,a}\right)} \cdot \frac{P\left( \mathcal{O}_{s,a} \right)}{P\left( \mathbf{h}^+, \mathbf{h}^-, \tau \right)} \\
    &= P\left(\mathcal{O}_{s,a} | \mathbf{h}^+, \mathbf{h}^-\right) P\left(\mathcal{O}_{s,a} | \tau \right) \frac{1}{P\left( \mathcal{O}_{s,a} \right)} \frac{P\left( \mathbf{h}^+, \mathbf{h}^- \right) P\left( \tau \right)}{P\left( \mathbf{h}^+, \mathbf{h}^-, \tau \right)} \\
    &\propto P\left(\mathcal{O}_{s,a} | \mathbf{h}^+, \mathbf{h}^-\right) P\left(\mathcal{O}_{s,a} | \tau \right) \frac{1}{P\left( \mathcal{O}_{s,a} \right)}.
\end{split}
\end{equation}

This final proportional expression highlights how the posterior probability of the optimality variable given both human feedback and trajectory observations can be decomposed into the product of two posterior terms, each conditioned on a different source of information, and adjusted by the prior probability of the optimality variable.

\section{Appendix C: Evaluation Details}
Here we describe our environmental setup for reproducibility of the experiments described in our paper. Across all environments, we compare our entropy-based active querying and crowd feedback aggregation approach against (1) the baseline approach (no feedback) and (2) random active learning (feedback randomly sampled from trajectories). We train an oracle policy as a proxy for an expert human trainer using standard reinforcement learning on the target environment.  Full implementations are provided in our code appendix.

\subsection{Gridworld Domains}
We train both an (1) oracle as a synthetic human trainer and (2) an agent trained with the additional feedback generated from (1). Table~\ref{tab:gridworld-params} details our gridworld agent and oracle training parameters. Oracle training uses Greedy Tabular Q-Learning over a single trial.

\begin{table}[h]
\centering
\small
\renewcommand{\arraystretch}{1.2}
\begin{tabular}{|l|l|}
\hline
\multicolumn{2}{|c|}{\textbf{GridWorld Experimental Parameters}} \\
\hline
\multicolumn{2}{|c|}{\textbf{Agent Training Parameters}} \\
\hline
Learning Rate ($\alpha$) & 0.05 \\
Discount Factor ($\gamma$) & 0.9 \\
Boltzmann Temperature ($\tau_b$) & 1.5 \\
Prior of $C_l$ & $\alpha_l = 90$, $\beta_l = 10$ for all trainers $l$ \\
Number of Trainers & 4 \\
Trainer Consistency Levels ($C_l$) & $[0.9,\ 0.8,\ 0.6,\ 0.3]$ \\
Episodes per Trial & 1000 \\
Number of Trials & 50 \\
Max Steps per Episode & 500 \\
\hline
\multicolumn{2}{|c|}{\textbf{Oracle Training Parameters}} \\
\hline
Learning Rate ($\alpha$) & 0.05 \\
Discount Factor ($\gamma$) & 0.9 \\
\hline
\rowcolor[gray]{0.9}
\multicolumn{2}{|l|}{\textbf{PACMAN}} \\
Episodes & 50000 \\
Max Steps per Episode & 500 \\
\hline
\rowcolor[gray]{0.9}
\multicolumn{2}{|l|}{\textbf{Taxi}} \\
Episodes & 50000 \\
Max Steps per Episode & 500 \\
\hline
\rowcolor[gray]{0.9}
\multicolumn{2}{|l|}{\textbf{Frozen Lake}} \\
Episodes & 5000 \\
Max Steps per Episode & 1000 \\
\hline
\end{tabular}
\caption{GridWorld experimental parameters for agent and oracle training.}
\label{tab:gridworld-params}
\end{table}

\begin{figure}
    \centering
    \begin{subfigure}[t]{0.4\textwidth}
        \centering
        \includegraphics[height=2.1in]{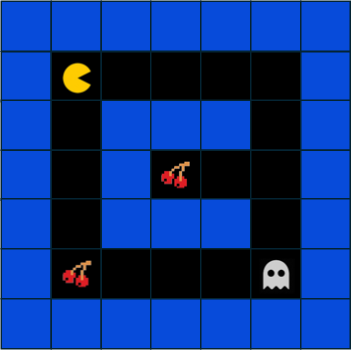}
        \caption{Default PACMAN Environment}
        \label{fig:default_pacman_env}
    \end{subfigure}
    ~
    \begin{subfigure}[t]{0.4\textwidth}
        \centering
        \includegraphics[height=2.3in]{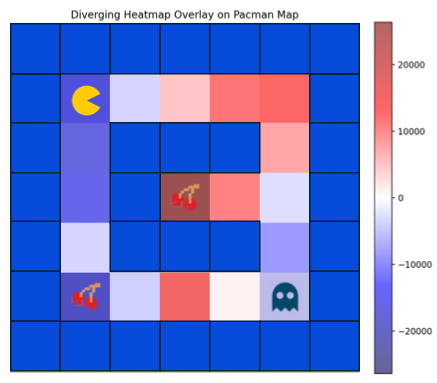}
        \caption{Feedback density comparison between random and active learning. Higher values indicate more entropy-based feedback, lower values indicate more random feedback.}
        \label{fig:pacman_heatmap}
    \end{subfigure}
    \caption{PACMAN Environment; (a) shows the starting positions of PACMAN and the ghost, as well as the two pellets which are to be collected to complete the game, (b) shows the different in feedback locations from random active learning and entropy-based active learning}
    \label{fig:pacman_envs}
\end{figure}

\subsubsection{PACMAN}
Our (5x5) PACMAN grid is illustrated in Figure~\ref{fig:default_pacman_env}. Figure~\ref{fig:pacman_heatmap} shows feedback generation aggregated by location and compares these aggregations between randomly generated feedback and that generated by active learning. We see a strong concentration of feedback around the second piece of fruit and in the top left of the map.

\subsubsection{Taxi}
\begin{figure}
    \centering
    \includegraphics[height=4in]{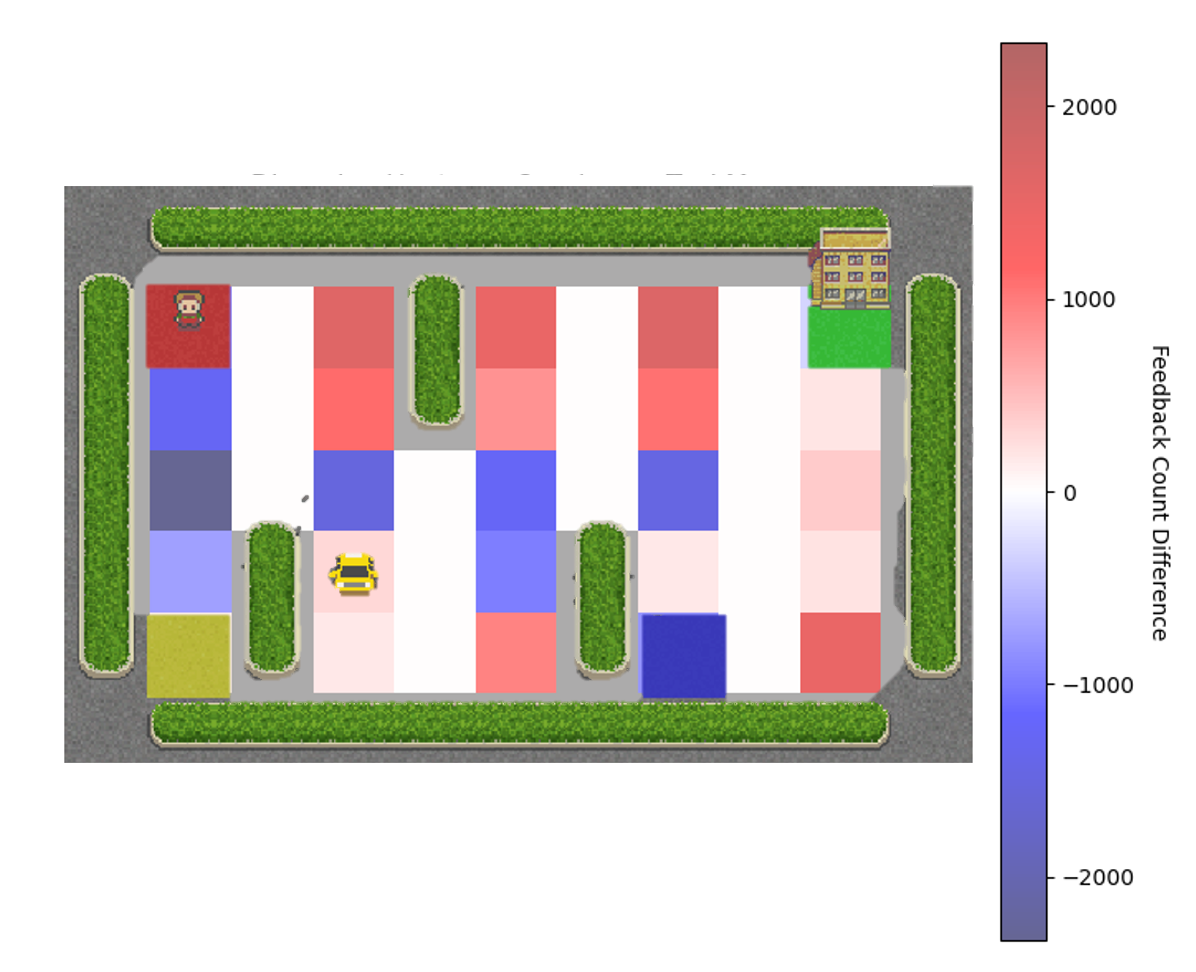}
    \caption{Taxi Feedback Comparison; comparing feedback sampled per location for random feedback selection and entropy-based feedback. The white ranks vertically adjacent to hedges should be ignored as they are not real states.}
    \label{fig:taxi_heatmap}
\end{figure}

Complementary evaluation featuring random starting positions and destinations and a larger map. Default parameters from the gymnasium library were used without modification
Farama Gymnasium suite \cite{towers2024gymnasiumstandardinterfacereinforcement}. Figure~\ref{fig:taxi_heatmap} shows the comparison of feedback gathered by random sampling and active learning. It appears that in this case (where random sampling and entropy-based sampling performed similarly) that entropy-based feedback seems to explore more. It could be that we are falsely attributing entropy earlier in learning when we have very little information. In the Taxi environment, individual states are less critical as the start and end position move between the red, blue, yellow and green tiles. Also there is no ghost chasing and forcing the agent into specific areas. This discrepancy in PACMAN and Taxi led us to investigate literature on critical states \cite{kumar2022preferofflinereinforcementlearning,goyal2023infobottransferexplorationinformation,liu2023learningidentifycriticalstates} and utilise Frozen Lake to create environments with varying numbers of bottlenecks. Further research is needed to align these theories and provide a comprehensive assessment.

\subsubsection{Frozen Lake}
Customised versions of the Taxi environment from the Farama Gymnasium suite \cite{towers2024gymnasiumstandardinterfacereinforcement}, designed to investigate the relationship between task difficulty and performance of our active learning method. The text version of our four maps are provided in Table~\ref{tab:frozenlake_variants}.

\begin{table}[H]
    \centering
    \renewcommand{\arraystretch}{1.2}
    \setlength{\tabcolsep}{12pt}
    \begin{tabular}{>{\centering\arraybackslash}p{0.4\textwidth} >{\centering\arraybackslash}p{0.4\textwidth}}
        \hline \hline
        \textbf{(0) Random, Multiple Solutions} & \textbf{(1) Easy} \\
        \hline      
        \begin{tabular}{c}
            \texttt{SFFHFFFF} \\
            \texttt{FHFFFHFG} \\
            \texttt{FFHHFFFH} \\
            \texttt{FFFFFHFF} \\
            \texttt{HFHFFHFF} \\
        \end{tabular}
        &
        \begin{tabular}{c}
            \texttt{SFFHHHFF} \\
            \texttt{FFFFHHFG} \\
            \texttt{FFFFFFFF} \\
            \texttt{FFFFHHFF} \\
            \texttt{FFFHHHFF} \\
        \end{tabular} \\
        \hline \hline
        \textbf{(2) Medium} & \textbf{(3) Hard (Maze)} \\
        \hline
        \begin{tabular}{c}
            \texttt{SFHFFHHF} \\
            \texttt{FFHFFHFG} \\
            \texttt{FFHFFFFF} \\
            \texttt{FFHFFHFF} \\
            \texttt{FFFFFHHF} \\
        \end{tabular} 
        &
        \begin{tabular}{c}
            \texttt{SFHFFFFH} \\
            \texttt{HFHHFHHG} \\
            \texttt{HFHFFFHF} \\
            \texttt{HFHFHFHF} \\
            \texttt{FFFFHFFF} \\
        \end{tabular} \\
    \end{tabular}
    \caption{Frozen Lake Environment Variants}
    \label{tab:frozenlake_variants}
\end{table}

\begin{figure}
    \centering
    \begin{subfigure}[t]{0.31\textwidth}
        \centering
        \includegraphics[height=3in]{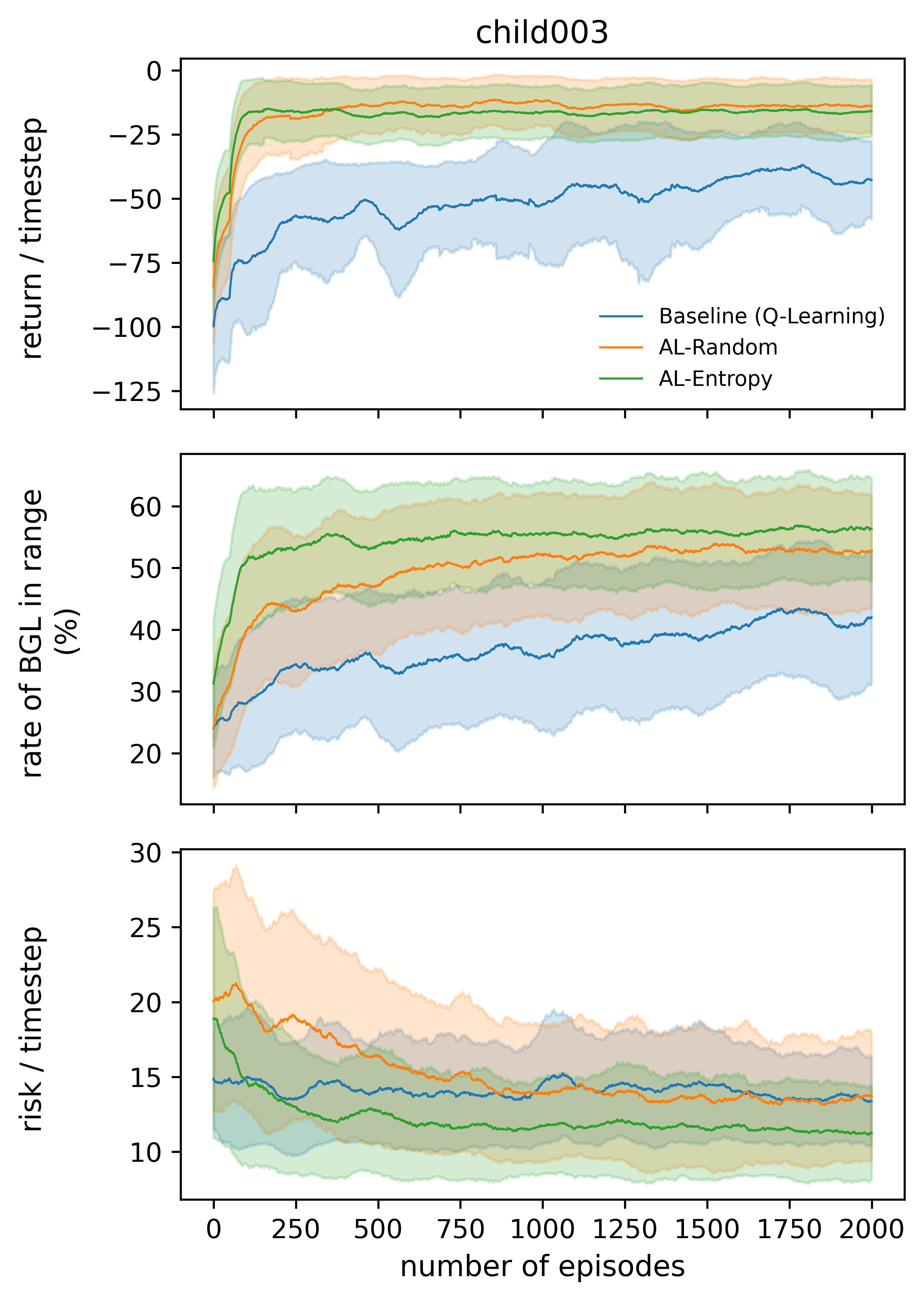}
        \caption{Child \#003}
    \end{subfigure}
    ~
    \begin{subfigure}[t]{0.31\textwidth}
        \centering
        \includegraphics[height=3in]{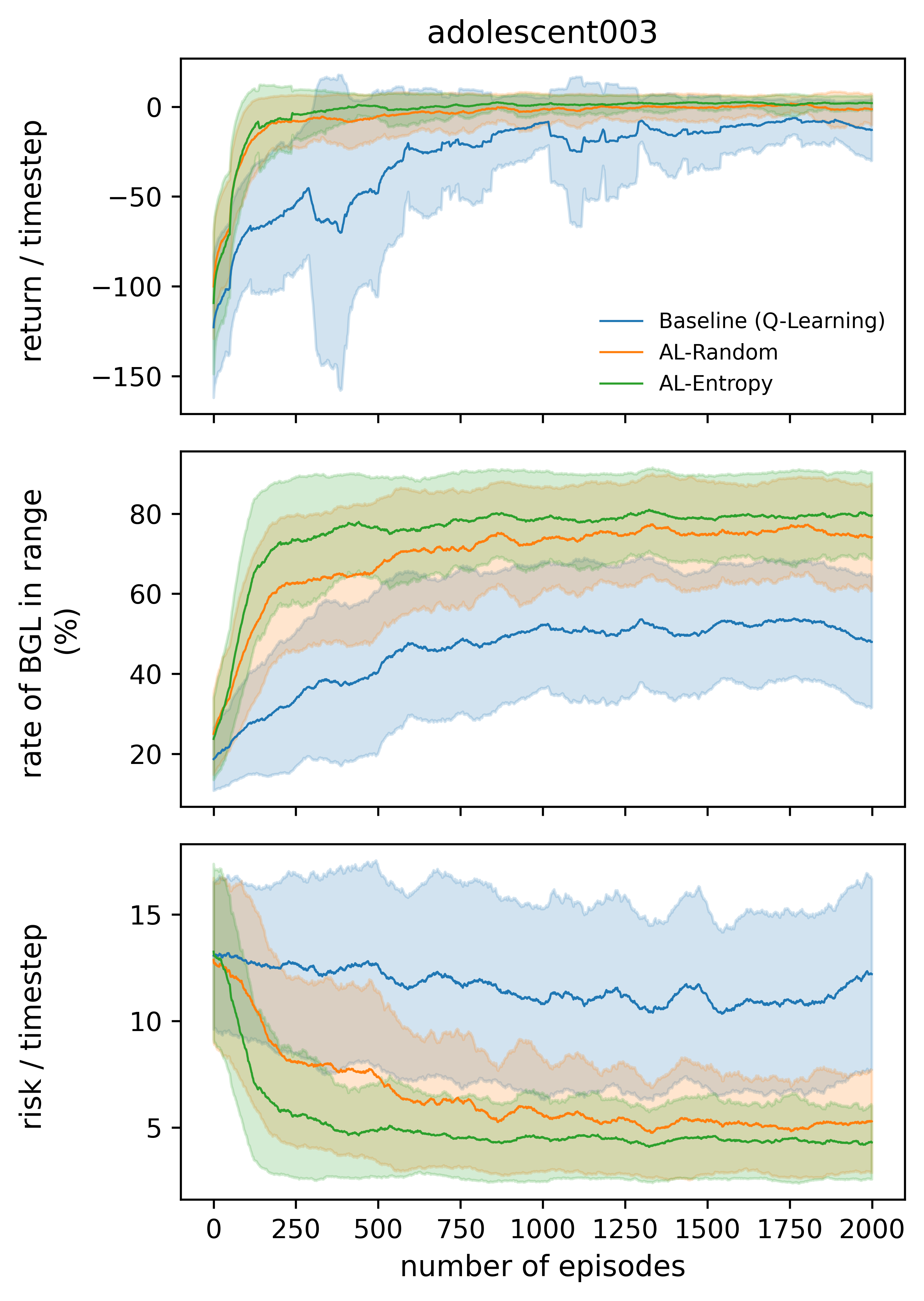}
        \caption{Adolescent \#003}
    \end{subfigure}
    ~
    \begin{subfigure}[t]{0.31\textwidth}
        \centering
        \includegraphics[height=3in]{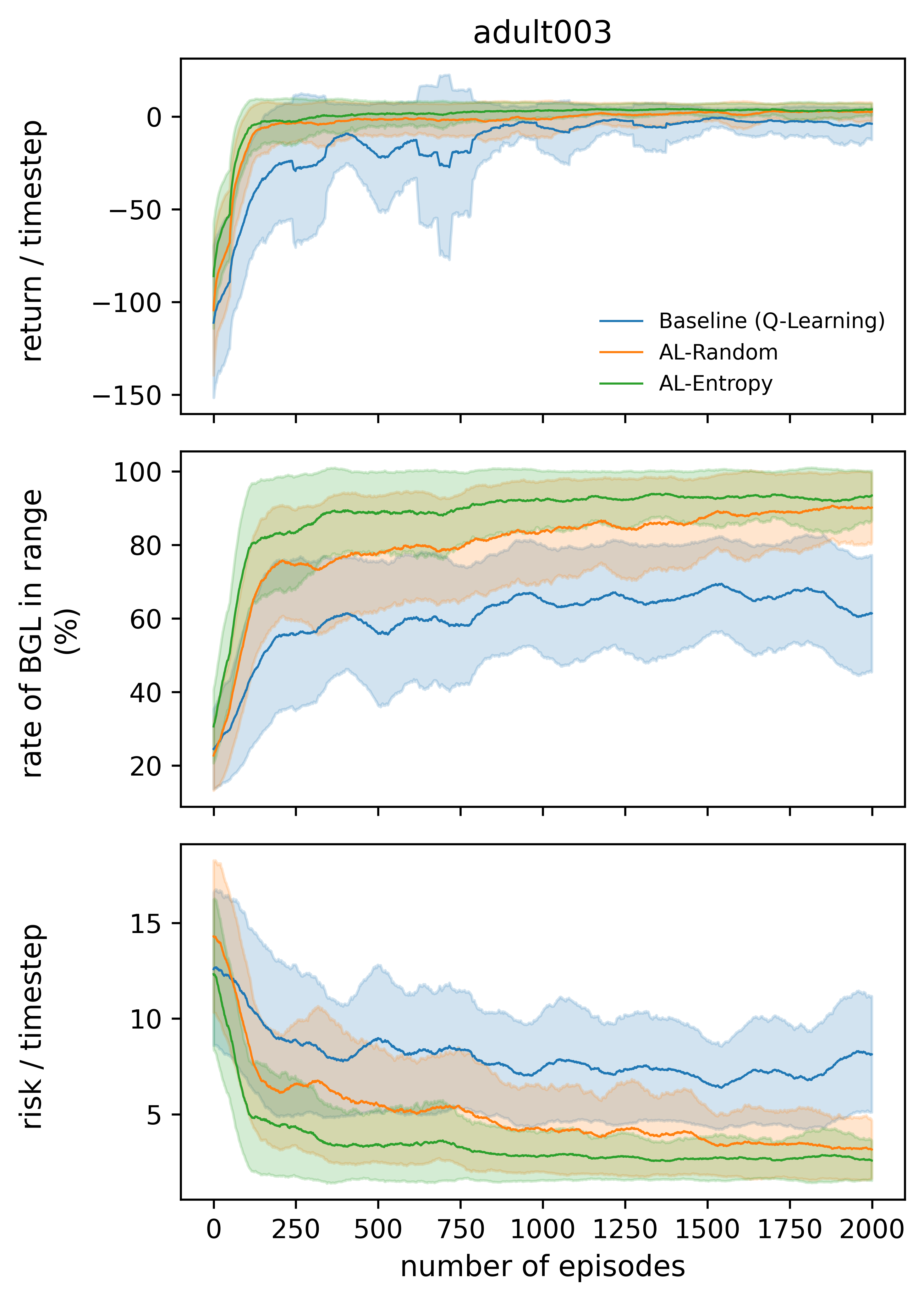}
        \caption{Adult \#003}
    \end{subfigure}
    
    \caption{Simulation results for three person profiles with five learning trials for 2,000 episodes.}
    \label{fig:diabetes_result_plots}
\end{figure}

\subsection{Type 1 Diabetes BGL Control}
We use an UVA/Padova simulator (FDA-approved Type 1 Diabetes simulator)~\cite{DallaMan2014}, which takes meal and insulin injection information, then outputs a blood glucose level (BGL) as a CGM reading at each time step. The algorithms (agents) receive the meal, insulin and blood glucose level information and decide the amount of insulin taken in the next time step. We simulate the algorithms together with the Type 1 Diabetes simulator and evaluate how well the blood glucose levels are managed.

\subsubsection{Environment}
Our simulator is based on an open-source implementation of UVA/Padova simulator~\cite{Xie2018}, and we modified the state and action spaces as follows.
For the action space, the insulin dose is discretised into six levels: 0, 20, 40, 80, 160, and 320 times the individual's basal insulin dose.

The state space is composed of the current \ac{BGL} reading from \ac{CGM}, the amount of carbohydrate intake,  insulin on board (IOB) and time of day (ToD). Each of these elements is discretised as the following ranges:

\begin{itemize}[noitemsep]
    \item \textbf{BGL}: [0, 70), [70, 90), [90, 110), [110, 180), [180, 300), [300, $\infty$] [mg/dL]
    \item \textbf{Carbs}: [0, 2), [2, 10), [10, 20), [40, 60), [60, 80), [80, $\infty$] [g]
    \item \textbf{IOB}: [0, 1), [1, 2), [2, 4), [4, $\infty$] [units]
    \item \textbf{ToD}: [0, 6), [6, 12), [12, 18), [18, 24) [hour]
\end{itemize}

The IOB is the amount of active insulin remaining in the body, which is computed by 25\% reduction of the amount of insulin every hour (assuming the insulin is active for four hours in the body). i.e. one insulin unit injected 1 hour ago and two units injected 3 hours ago:
\[
\text{IOB} = 1 \times 0.75 + 2 \times 0.25 = 1.25
\]
\subsubsection{Experimental Setup}
We employ Q-Learning as the underlying \ac{RL} algorithm combined with Boltzmann exploration policy. To simulate human feedback, we employ BBController as the oracle. The experiments are carried out 5 learning trials (each with different random seeds) for 2,000 episodes. We evaluate three virtual profiles: child\#003, adolescent\#003 and adult\#003.

For this environment, we simulate 5 trainers with consistency levels of 0.9, 0.8, 0.7, 0.6, 0.3 who provide the feedback 10\% of the time on average. Other hyper-parameters are set as follows:
\begin{itemize}[noitemsep]
    \item Learning Rate: $\alpha = 0.05$
    \item Discount Factor: $\gamma = 0.98$
    \item Boltzmann Exploration Temperature: $\tau_{b} = 10.0$
    \item Prior of $C_l$: $\alpha_l = 90$, $\beta_l = 10$ for all trainers $l$
    \item Base Variance of Q estimation: $\sigma^2_{base} = 5000$
    \item Max Steps per Episode: $480$
    \item CGM Hardware: Dexcom
    \item Insulin Pump: Insulet
\end{itemize}

\subsubsection{Supplemental Results}
Figure~\ref{fig:diabetes_result_plots} presents the simulation results for three virtual profiles: child\#003, adolescent\#003 and adult\#003. Each simulation comprises five learning trials over 2,000 episodes. The figure is organised into three rows, presenting the following from top to bottom: the return (sum of reward) per time step, the percentage of time that BGL stays in the target range (70 - 180 mg/dL) and the averaged risk index per time step. The x-axis for all plots represents the number of episodes.

The figures show that our approach (AL-Entropy) outperforms other baselines (Baseline and AL-Random), which indicates the effectiveness of our entropy-based active feedback algorithm.





\ifreproStandalone
\clearpage
\bibliography{aaai2026}
\end{document}
\fi

\end{document}